\documentclass{article}
\usepackage{graphicx} 
\usepackage{subcaption}
\usepackage{multirow}
\usepackage{float}
\usepackage{xurl}
\usepackage[table]{xcolor}
\usepackage{geometry}
\usepackage{indentfirst}
\usepackage{comment}
\usepackage{authblk}
\usepackage{makecell}

\title{The Impact of Quantization and Pruning on Deep Reinforcement Learning Models}
\author[1]{Heng Lu}
\author[2,3]{Mehdi Alemi}
\author[1]{Reza Rawassizadeh}
\affil[1]{Department of Computer Science at Metropolitan College, Boston University, Boston, MA, USA}
\affil[2]{Department of Orthopaedic Surgery, Harvard Medical School, Boston, MA, USA.}
\affil[3]{Training Services, MathWorks, Natick, MA, USA.}

\date{}

\begin{document}
    
\maketitle
\begin{abstract}
    Deep reinforcement learning (DRL) has achieved remarkable success across various domains, such as video games, robotics, and, recently, large language models. However, the computational costs and memory requirements of DRL models often limit their deployment in resource-constrained environments. The challenge underscores the urgent need to explore neural network compression methods to make RDL models more practical and broadly applicable. Our study investigates the impact of two prominent compression methods, \emph{quantization} and \emph{pruning} on DRL models. We examine how these techniques influence four performance factors: average return, memory, inference time, and battery utilization across various DRL algorithms and environments. Despite the decrease in model size, we identify that these compression techniques generally do not improve the energy efficiency of DRL models, but the model size decreases. We provide insights into the trade-offs between model compression and DRL performance, offering guidelines for deploying efficient DRL models in resource-constrained settings. 
\end{abstract}
\section{Introduction and Background}
Reinforcement learning has been applied in many fields, including robotics, video games, and recently Reinforcement Learning with Human Feedback (RLHF) \cite{alphago, games, dreamer, rlhf1, control1}, has become common in large language models. RLHF methods mitigate biases inherent in language models themselves \cite{rlhf1,llama2,codellama}. Reinforcement learning models that address real-world problems predominantly utilize continuous models based on neural network architecture, known as Deep Reinforcement Learning (DRL). 

DRL methods typically involve a world model, agents interacting with the world, and a reward function that evaluates the effectiveness of actions based on the agent's policy towards predefined objectives \cite{rl}. Depending on whether the algorithm learns a specific world model, DRL algorithms are categorized into model-based DRL algorithms and model-free DRL algorithms \cite{mbrl}. Model-free DRL methods generally fall into three main categories: deep Q-learning methods \cite{dqn,cql,sql}, policy gradient methods \cite{reinforce,trpo, ppo}, and actor-critic methods \cite{a2c,ddpg,td3,sac-1}. Unlike model-based algorithms, model-free approaches circumvent model bias and offer greater generalizability, which contributes to their popularity in RLHF applications \cite{rlhf1,rlhf2}. 

Neural networks, which are the backbone of DRL methods, are associated with high computational costs and, therefore, resource intensive. Recently, there has been a significant increase in the energy and water consumption of artificial intelligence (AI) data centers \footnote{\url{https://www.theatlantic.com/technology/archive/2024/03/ai-water-climate-microsoft/677602}}\footnote{\url{https://www.oregonlive.com/silicon-forest/2022/12/googles-water-use-is-soaring-in-the-dalles-records-show-with-two-more-data-centers-to-come.html}}\footnote{\url{https://www.bloomberg.com/news/articles/2023-07-26/thames-water-considers-restricting-flow-to-london-data-centers}}\footnote{\url{https://www.washingtonpost.com/business/2024/03/07/ai-data-centers-power}}. This trend has led to several research studies \cite{Luccioni2023, deVries2023, Li2023} investigating the resource costs of recent advances in AI. Reducing the energy utilization of DRL has become a crucial need.

Additionally, many systems that benefit from reinforcement learning operate on battery-powered devices, such as extended reality devices and mobile robots. As the size of these devices decreases, their computational capabilities also diminish \cite{wearable}. One common approach to reducing the computational costs of neural network models is compressing them via pruning and quantization\cite{dc,pr}. Network compression methods have been widely applied in computer vision \cite{cvQuant1, cvQuant2} and large language models \cite{gptq, llmQuant1, llmQuant2} to improve inference times and reduce memory requirement with minimal compromise to accuracy. Compression enables advanced DRL models to be deployed in robots with low latency and high energy efficiency under constrained resources. Despite its promise, neural network compression in DRL models has received less research attention \cite{ActorQ, FIXAR} compared to other fields, such as computer vision and natural language processing.

Several general approaches have been proposed for neural network compression method quantization and pruning \cite{OPQ, quant}. As for quantization, DRL researchers might be more familiar with vector quantization, which aims to discretize continuous space into a discrete vector set to reduce dimensions \cite{VQ, DVQ, VQMP, AQua}. However, in this work, we specifically apply neural network quantization, which focuses on converting float32 format weights and biases into smaller-scale numbers such as int8 or 4-bits rather than vector quantization. Pruning methods, on the other hand, involve removing neurons deemed least important based on criteria such as weight or activation value \cite{QPrune, pops, relp, DSP, DoubleSparse}. 

In addition to conventional compression approaches \cite{quant, dc, pr}, there are promising DRL-specific compression approaches \cite{ActorQ, VQ, QPrune, DSP}.  AQuaDem \cite{AQua} discretizes the action space and learns a discrete set of actions for each state $s$ using behavior cloning from expert demonstrations \cite{dagger}. The adaptive state aggregation algorithm\cite{VQ} adaptively discretizes the state space based on Bregman divergence, enabling distinct partitions of the state space. Another group of methods focuses on scalar quantization that reduces the numeric precision of values \cite{AccDRL, QPrune, FIXAR, ActorQ}. NNC-DRL\cite{AccDRL} accelerates the DRL training process by speeding up prediction based on GA3C\cite{GA3C} and employs policy distillation to compress the behavior policy network prediction with minimal performance degradation. FIXAR \cite{FIXAR} proposes quantization-aware training in fixed points to reduce model size without significant accuracy loss. ActorQ \cite{ActorQ} introduces an int8 quantized policy block for rollouts within a traditional distributed RL training loop. PoPS\cite{pops} accelerates model speed by initially training a sparse network from a large-scale teacher network through iterative policy pruning, then compacting it into a dense network with minimal performance loss. UVNQ\cite{QPrune} integrates sparse variational dropout\cite{SVD} with quantization, adjusting the dropout rate to enhance quantization awareness. Dynamic Structured Pruning\cite{DSP} enhances DRL training by applying a neuron-importance group sparse regularizer and dynamically pruning insignificant neurons based on a threshold. Double Sparse Deep Reinforcement Learning \cite{DoubleSparse} uses multilayer sparse-coding structural network with a nonconvex log regularizer to enforce sparsity while maintaining performance.

In this work, we apply common neural network compression methods, including common quantization and pruning, to five popular deep reinforcement learning models (TRPO\cite{trpo}, PPO\cite{ppo}, DDPG\cite{ddpg}, TD3\cite{td3}, and SAC\cite{sac-1}). We then measure the performance of these algorithms post-compression using metrics such as average return, inference time, and energy usage. In particular, we apply $L_1$ and $L_2$ pruning techniques to these models. For quantization, we utilize \emph{int8} quantization and apply (i) post-training dynamic quantization, (ii) post-training static quantization, and (iii) quantization aware training across the listed models. 

To our knowledge, this study represents the first comprehensive evaluation of the effects of pruning and quantization across a range of deep reinforcement learning models. Our experiments and findings offer valuable insights to researchers and developers, assisting them in making informed decisions when choosing between quantization or pruning methods for DRL models. Another prevalent approach for compressing neural network is knowledge distillation \cite{gou}, but due to its model or application-specific nature (e.g., image classification), we did not include knowledge distillation in our experimental setup.

\section{Methods} \label{method}
We have applied two types of neural network compression techniques —pruning and quantization— across five prominent DRL models: TRPO\cite{trpo}, PPO\cite{ppo}, DDPG\cite{ddpg}, TD3\cite{td3}, and SAC\cite{sac-1}. This section outlines our quantization methods, followed by our pruning methods.

\subsection{Quantization}

We applied linear quantization across all models, where the relationship between the original input $r$ and its quantized version $q$ is defined as $r = S(q+Z)$. Here, $Z$ represents the zero point in the quantization space, and the scaling factor $S$ maps floating-point numbers to the quantization space. For Post-Training Dynamic Quantization (PTDQ) and Post-Training Static Quantization (PTSQ), we computed $S$ and $Z$ for activations exclusively. In PTSQ, first, baseline models go through a calibration process to compute these quantization parameters and then the models make inferences based on the fixed quantization parameters. In PTDQ, the quantization parameters are computed dynamically. In Quantization-Aware Training (QAT), baseline models are pseudo-quantized during training, meaning computations are conducted in floating-point precision but rounded to integer values to simulate quantization. Subsequently, the original models are converted into quantized versions, and the quantization parameters are stabilized.


\subsection{Pruning}
Neural network pruning typically involves removing neurons within layers, and dependencies can exist where pruning in one layer affects subsequent related layers.  The DepGraph approach we employed \cite{dep}, addresses these dependencies by grouping layers based on their inter-dependencies rather than manually resolving dependencies.

Conceptually, one might consider constructing a grouping matrix $G\in R^{L\times L}$, where $G_{ij} = 1$ signifies a dependency between layer $i$ and layer $j$. However, due to the complexity arising from non-local relations, $G$ can not be easily constructed. Thus, dependency graph $D$ is proposed, which only contains the local dependency between adjacent layers and from which the grouping matrix can be reduced. These dependencies are categorized into two types: inter-layer dependencies, where the output of one layer $i$ connects to the input of another layer $j$, and intra-layer dependencies, such as within BatchNorm layers, where inputs and outputs share the same pruning scheme. 

After constructing the dependency graph and determining grouping parameters based on this graph, we utilized a norm-based importance score. However, directly summing importance scores across different layers can lead to meaningless results and potential divergence. Therefore, for a parameter $w$ in group $g$ with $K$ prune-able dimensions, a regularization term $R(g,k)$ is used in sparse training to select the optimal input variables, 
$
R(g,k) = \sum_{k=1}^K \gamma_k\cdot I_{g,k},
$
where $I_{g,k} = \sum_{w\in g} ||w[k]||_2^2$ is the importance for dimension $k$ in $L_2$ pruning and 
$
\gamma_k = 2^{\alpha(I_g^{max}-I_{g,k})/(I_g^{max}-I_g^{min})}.
$

\section{Experiments}\label{exp}
\subsection{Experimental Settings}
Our experiments are structured into two main components: quantization and pruning of DRL algorithms. We evaluated the performance of TRPO\cite{trpo}, PPO\cite{ppo}, DDPG\cite{ddpg}, TD3\cite{td3}, and SAC\cite{sac-1} across five Gymnasium\cite{gymnasium} (formerly OpenAI Gym\cite{gym}) Mujoco environments including: HalfCheetah, HumanoidStandup, Ant, Humanoid, and Hopper. These models are trained using Gymnasium (formerly OpenAI Gym) environments. 

To ensure consistency among our reported results, each experiment has been repeated at least 10 times in the same configuration. 

\textbf{Quantization and Pruning Libraries:}
The implementations of quantization and pruning in neural network libraries are not as mature as other functionalities. For instance, in pyTorch pruning does not remove neurons but merely masks them. To ensure the reliability of our experiments, we evaluated various quantization and pruning libraries and selected those that offer the highest accuracy and resource efficiency. 

Therefore, to implement pruning, we explored PyTorch\footnote{\url{https://pyTorch.org}} and Torch-pruning. In our experiment, Torch-pruning\footnote{\url{https://github.com/VainF/Torch-Pruning}}, integrated with DepGraph \cite{dep}, performed exceptionally well, and thus, we utilized it for pruning purposes. Regarding quantization, we experimented with Pytroch, TensorFlow, and ONNX Runtime\footnote{https://onnxruntime.ai}. Ultimately, we chose PyTorch for QAT, and ONNX Runtime for PTDQ and PTSQ.

\textbf{Hardware Settings:} Our hardware infrastructure included two NVidia RTX 4090 GPUs with 24GB of VRAM, 256GB of RAM, and an Intel Core i9 CPU running at 3.30 GHz. The operating system was Ubuntu 20.04 LTS, and we used CUDA Version 12.0 for GPU operations.

\subsection{Quantization}
To implement quantization, we experimented with three approaches: PTDQ, PTSQ and QAT. Quantization-aware training (QAT) involved initially training quantized models with an equivalent dataset size as the baseline models, followed by exporting them into ONNX runtime for comparative analysis.

\subsubsection{Average Return}
The impact of quantization on average return is reported in Table \ref{quantall}. The table underscores the variability of quantization outcomes across different environments and DRL models. For instance, QAT demonstrates its highest efficacy in HumanoidStandup environments, resulting in improved average returns across models except for PPO. The SAC algorithm generally benefits more from QAT, except in the Hopper environment, where its effectiveness is limited. Overall, PTDQ exhibits superior performance, while PTSQ consistently shows the lowest results. The observed performance discrepancies may stem from distribution shifts between data used for optimal path calculations and that utilized during the calibration phase, which are challenging to rectify due to the stochastic nature of the environment.
{
\begin{table}[H]
\centering
\small
\begin{tabular}{|l|l|l|l|l|l|}
\hline
\multicolumn{2}{|c|}{}  & Baseline & PTDQ & PTSQ & QAT \\ \hline

\multirow{5}{*}{TRPO} &HalfCheetah & 978.48 & \textbf{1004.65} & 909.52 & 287.94 \\ \cline{2-6}
                 &HumanoidStandup & 39444.51 & 37002.02 & 35779.53 &\textbf{52252.48} \\ \cline{2-6}
                  &Ant & 851.33 & 799.46 & \textbf{1055.39} &970.59\\ \cline{2-6}
                  &Humanoid & 74.49 & 75.09 & 74.98 &\textbf{178.36}\\ \cline{2-6}
                  &Hopper & 164.06 & \textbf{163.84} & 162.93 &7.49\\ \cline{2-6} \hline \hline
              \multirow{5}{*}{PPO} &HalfCheetah & 1542.98 & \textbf{1508.82} & 1482.4 & 432.21 \\ \cline{2-6}
                 &HumanoidStandup & 117138.8 & 126509.58 & \textbf{128155.96} &28270.98 \\ \cline{2-6}
                  &Ant & 1335.55 & 1493.93 & \textbf{1528.36} &963.21\\ \cline{2-6}
                  &Humanoid & 453.56 & 430.5 & \textbf{495.0} &295.75\\ \cline{2-6}
                  &Hopper & 8.33 & 9.41 & 19.28 &\textbf{92.32}\\ \cline{2-6} \hline \hline
              \multirow{5}{*}{DDPG} &HalfCheetah & 4475.62 & \textbf{4656.76} & 3801.11 & 937.53 \\ \cline{2-6}
                 &HumanoidStandup & 82747.99 & 82747.99 & 87346.75 &\textbf{115325.59} \\ \cline{2-6}
                  &Ant & 589.46 & 630.37 & \textbf{896.44} &540.7\\ \cline{2-6}
                  &Humanoid & 1544.76 & 314.51 & \textbf{433.15} &411.92\\ \cline{2-6}
                  &Hopper & 1419.69 & \textbf{1260.39} & 349.31 &996.61\\ \cline{2-6} \hline \hline
              \multirow{5}{*}{TD3} &HalfCheetah & 8333.37 & \textbf{5204.59} & 3915.44 & 4169.07 \\ \cline{2-6}
                 &HumanoidStandup & 77140.94 & 77140.94 & 78492.56 &\textbf{82211.28} \\ \cline{2-6}
                  &Ant & 3423.51 & \textbf{2789.51} & 2728.76 &1833.74\\ \cline{2-6}
                  &Humanoid & 5035.79 & \textbf{441.34} & 287.35 &80.67\\ \cline{2-6}
                  &Hopper & 3596.54 & \textbf{3532.45} & 2842.55 &1826.21\\ \cline{2-6} \hline\hline
              \multirow{5}{*}{SAC} &HalfCheetah & 10460.06 & 3104.33 & 1805.41 & \textbf{6163.8} \\ \cline{2-6}
                 &HumanoidStandup & 151015.38 & 109714.88 & 83898.23 &\textbf{151213.82} \\ \cline{2-6}
                  &Ant & 4021.96 & 2474.88 & 1144.59 &\textbf{3119.85}\\ \cline{2-6}
                  &Humanoid & 4287.29 & -84.51 & 23.4 &\textbf{311.68}\\ \cline{2-6}
                  &Hopper & 2539.05 & \textbf{3621.34} & 2525.01 &2998.79\\ \cline{2-6} \hline

\end{tabular}
\caption{Average returns of quantization for TRPO, PPO, DDPG, TD3, and SAC. The best quantized version for each DRL model on the specific environments is shown in bold. }
\label{quantall}
\end{table}
}

\subsubsection{Resource Utilization}
\begin{figure}[H]
    \centering
    \includegraphics[width=0.95\linewidth]{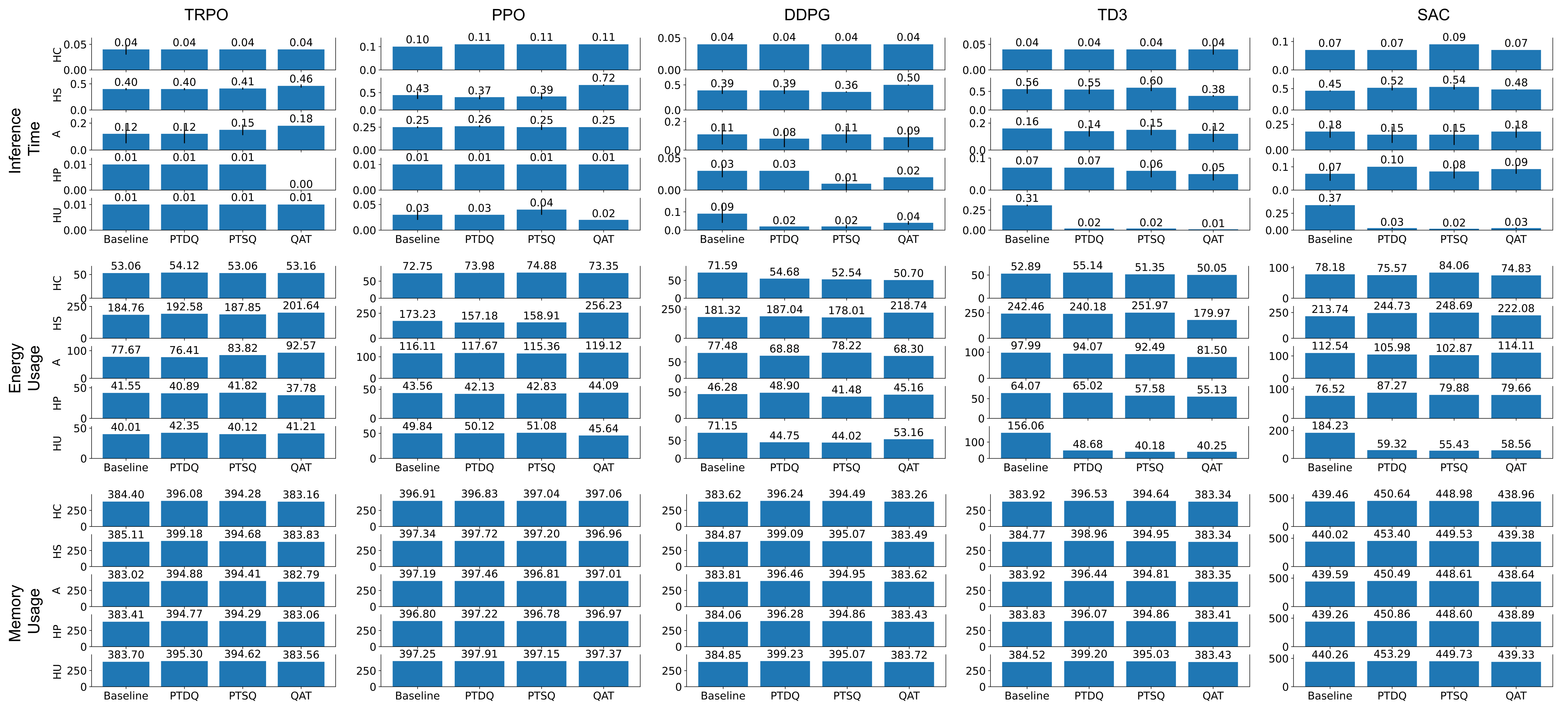}
    \caption{Inference time (in seconds), energy usage (in Joules) and memory utilization (in MegaByte) of quantization models.}
    \label{quant}
\end{figure}

To assess the impact of quantization on resource utilization, we conducted measurements and comparisons of memory usage, inference time, and energy consumption between baseline models and their quantized counterparts. Figure \ref{quant} illustrates the differences observed in inference time and energy usage between baseline and quantized models.

\subsection{Pruning}
To implement pruning, we utilized the torch-pruning package \footnote{https://github.com/VainF/Torch-Pruning} for all our experiments. Each baseline model underwent $L_1$ and $L_2$ pruning, with various pruning percentages ranging from $5\%$ to $70\%$. In particular, experimented pruning percentages are as follows: \{$5\%$,$10\%$,$15\%$,$20\%$,$25\%$,$30\%$,$35\%$,$40\%$,$45\%$,$50\%$,$55\%$,$60\%$,$65\%$,$70\%$\}.

\begin{table}[H]
\centering

\begin{tabular}{|l|l|l|l|l|l|}
\hline
\multicolumn{2}{|c|}{}   & Baseline & Pruned & \makecell{$L_1$ Pruning\\  Percentage} & \makecell{$L_2$ Pruning\\  Percentage} \\ \hline

\multirow{5}{*}{TRPO} &HalfCheetah & 1003.37 & 905.94 & 0.05 & 0.05 \\ \cline{2-6}
                 &HumanoidStandup & 35281.42 & 42314.62 & 0.55&0.6 \\ \cline{2-6}
                  &Ant & 768.97 & 979.42 & 0.7&0.7  \\ \cline{2-6}
                  &Humanoid & 74.49 & 79.81 & 0.25&0.35  \\ \cline{2-6}
                  &Hopper & 164.06 & 163.33 & 0.15&0.45 \\ \cline{2-6} \hline
              \multirow{5}{*}{PPO} &HalfCheetah & 1529.18 & 1453.9 & 0.1&0.05 \\ \cline{2-6}
                 &HumanoidStandup & 128722.71 & 117239.65 & 0.1&0.1 \\ \cline{2-6}
                  &Ant & 1563.54 & 263.37 & 0.05&0.05 \\ \cline{2-6}
                  &Humanoid & 474.86 & 417.99 & 0.1&0.15 \\ \cline{2-6}
                  &Hopper & 21.06 & 7.6 & 0.3&0.3 \\ \cline{2-6} \hline
              \multirow{5}{*}{DDPG} &HalfCheetah & 5104.06 & 4026.46 & 0.05&0.05 \\ \cline{2-6}
                 &HumanoidStandup & 82747.99 & 83513.05 & 0.7&0.7 \\ \cline{2-6}
                 &Ant & 935.72 & 761.55 & 0.1&0.05 \\ \cline{2-6}
                  &Humanoid & 1659.59 & 1059.68 & 0.05&0.05 \\ \cline{2-6}
                  &Hopper & 1574.66 & 1423.7 & 0.05&0.05 \\ \cline{2-6} \hline
              \multirow{5}{*}{TD3} &HalfCheetah & 8298.95 & 6535.42 & 0.05&0.05 \\ \cline{2-6}
                 &HumanoidStandup & 77140.94 & 97061.37 & 0.7&0.7 \\ \cline{2-6}
                  &Ant & 3381.72 & 2564.85 & 0.05&0.05 \\ \cline{2-6}
                  &Humanoid & 5040.01 & 5046.25 & 0.1&0.1 \\ \cline{2-6}
                  &Hopper & 3593.0 & 3589.47 & 0.05&0.05 \\ \cline{2-6} \hline
              \multirow{5}{*}{SAC} &HalfCheetah & 10467.97 & 10531.88 & 0.05&0.05 \\ \cline{2-6}
                 &HumanoidStandup & 136900.13 & 137574.71 & 0.7&0.7 \\ \cline{2-6}
                  &Ant & 3499.52 & 282.28 & 0.05&0.05 \\ \cline{2-6}
                  &Humanoid & 4251.2 & 3549.46 & 0.25&0.35 \\ \cline{2-6}
                  &Hopper & 2549.96 & 2412.92 & 0.05&0.2 \\ \cline{2-6} \hline

\end{tabular}
\caption{Average returns of pruning results for TRPO, PPO, DDPG, TD3, and SAC}
\label{prune:l1}
\end{table}

\begin{figure}[H]
    \centering
    \includegraphics[width=\linewidth]{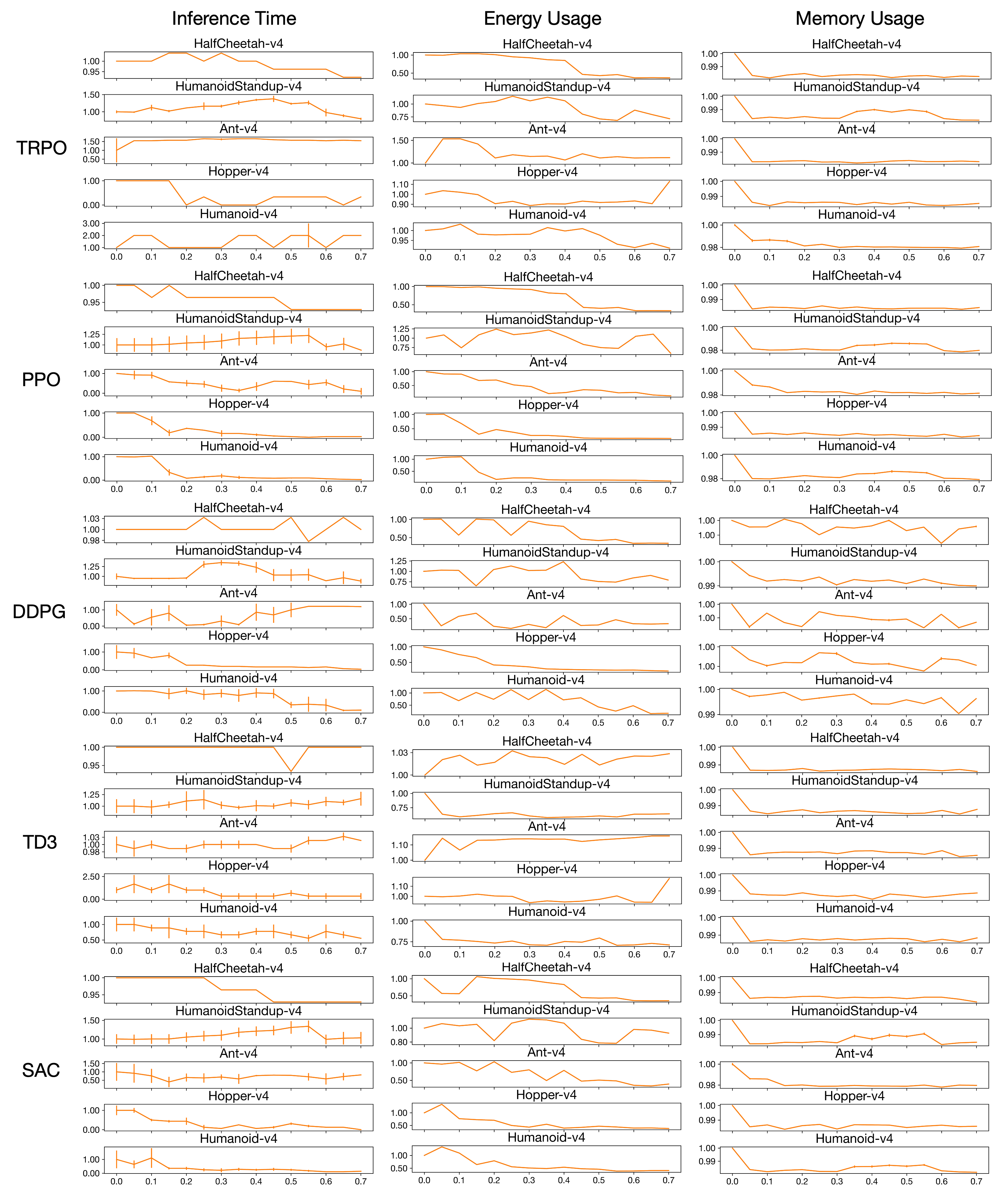}
    \caption{Inference time, energy usage and RAM of $L_1$ models, scaled by baseline models}
    \label{l1}
\end{figure}

\begin{figure}[H]
    \centering
    \includegraphics[width=\linewidth]{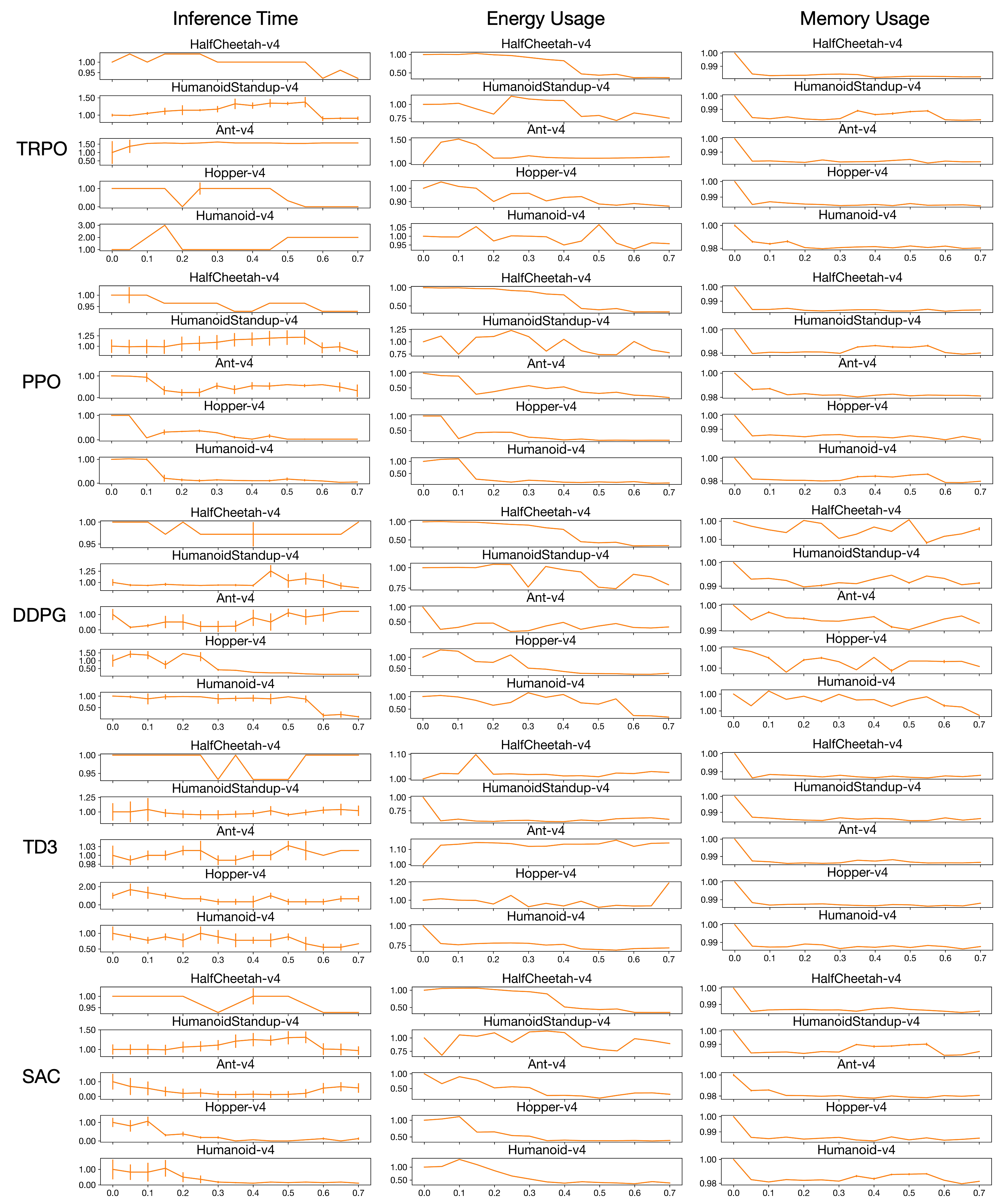}
    \caption{Inference time (in seconds), energy usage (in Joules) and memory utilization (in Megabytes) of $L_2$ models, scaled by baseline models.}
    \label{l2}
\end{figure}

The optimal pruning method for each baseline model was determined based on earning at least $90\%$ average return of the corresponding baseline model while achieving the highest possible pruning percentage. The results of pruning experiments are presented in Table \ref{prune:l1} and summarized comprehensively in Table \ref{models:pruning}. In Figure \ref{l1} and Figure \ref{l2}, we present the effect of $L_1$ and $L_2$ pruning on the inference speed, energy usage, and memory usage. In these figures, we scaled the data according to the baseline.
\begin{table}[H]
\centering
\begin{tabular}{|c|c|c|c|c|c|}
\hline
	& TRPO & PPO & DDPG & TD3 & SAC \\ \hline
HalfCheetah & $L_1$ 5\% &  $L_1$ 10\% &  $L_2$ 5\% &  $L_2$ 5\% &  $L_1$ 5\% \\ \hline
HumanoidStandup & $L_2$ 55\% &  $L_1$ 5\% &  $L_1$ 70\% &  $L_1$ 70\% &  $L_2$ 70\% \\ \hline
Ant & $L_2$ 70\% &  $L2$ 5\% &  $L_1$ 10\% & $L_2$ 5\% &  $L_2$ 25\% \\ \hline
Humanoid & $L_2$ 30\% & $L_1$ 5\% &  $L_2$ 5\% &  $L_1$ 10\% &  $L_2$ 25\% \\ \hline
Hopper & $L_2$ 40\% &  $L_2$ 30\% &  $L_1$ 5\% &  $L_2$ 5\% &  $L_2$ 20\% \\ \hline
\end{tabular}
\caption{Best pruning method for each environment and each model.}
\label{models:pruning}
\end{table}

\hspace{-4cm}

\section{Discussions and Findings}
In this work, we studied two pruning approaches and three quantization approaches on five platforms (HalfCheetah-v4, HumanoidStandup-v4, Ant-v4, Hopper-v4, and Humanoid-v4) used for experimenting reinforcement learning methods and five common DRL methods (TRPO, PPO, DDPG, TD3). To our knowledge, this is the largest study performed on compressing DRL methods, and we listed our findings in this section. These findings could be used as a guideline for further studies that try to compress DRL methods. 

\emph{Pruning and quantization do not improve the energy efficiency and memory usage of DRL models.} While pruning and quantization reduce model size (see Table \ref{models:pruning}), they do not necessarily enhance the energy efficiency of DRL models due to the maintained or increased average return. Energy consumption tends to decrease only when there is a significant drop in average return, prompting the agent to terminate early and requiring less computation. 

\emph{Despite reducing model size, quantization does not improve memory usage, and pruning yields only a negligible 1\% decrease in memory usage.} Results in Figure \ref{quant} present no changes in memory utilization in any platforms while applying quantization. Even PTDQ and PTSQ cause more memory utilization than the baseline method. This might be due to the overhead of the quantization library, and the way it is implemented is not optimized.

\textit{$L_2$ pruning is favored over $L_1$ pruning for most of DRL models.} Table \ref{prune:l1}-\ref{models:pruning} illustrates that the optimal pruning method varies based on the DRL algorithm and environmental complexity. Most environments, except for those trained with SAC on HalfCheetah, allow for substantial pruning without a notable decline in average return, while PPO models exhibit lower pruning thresholds. In instances where $L_1$ pruning outperforms $L_2$, the average return values remain closely aligned. Generally, a 10\% reduction in DRL model size through $L_2$ pruning is beneficial, although exceptions include PPO models applied to HalfCheetah environments.

\textit{PTDQ emerges as the superior quantization method for DRL algorithms, whereas PTSQ is not recommended.} As shown in Table \ref{quantall}, 40\% of our quantized models benefit from PTDQ, 36\% from QAT, and only 24\% from PTSQ. Our findings reveal that post-training dynamic quantization statistically outperforms the other methods, while post-training static quantization performs the worst, likely due to distribution shifts between the calibration data and the randomness that existed in RL environments.

\textit{The Lottery ticket hypothesis \cite{lottery} does not hold for DRL models.} 
The Lottery Ticket Hypothesis (LTH) in the context of neural networks suggests that within a large, randomly initialized network, there exists a smaller sub-network, typically around 10-20\% of the original size, that, when trained in isolation, can achieve performance comparable to the original large network. This idea has significant implications for model quantization and pruning, two techniques used to reduce the size and computational requirements of neural networks. However, based on the results demonstrated in Table \ref{prune:l1}  demonstrate significant performance drops in most models after 50\% pruning, contradicting the hypothesis's assertion that original network performance can persist even when pruned to less than 10\%-20\% of its original size. In particular, around 40\% of the models don't survive after more than 5\% pruning, but 80\% of the models don't survive after 50\%.

Our work has two limitations. First, by focusing on classical Mujoco environments with continuous action spaces, our work excludes discrete action spaces, which are common in video games or some decision-making scenarios. However, in these situations, task-specific methods might be employed for a satisfying performance, which adds additional complexities, and we might explore it in our future work. Moreover, we are limited to six simulated environments, this leaves an important aspect of real-world applicability unexplored. An ideal scenario is to experiment with this compression approach on a robot or drone in a real-world task and measure the differences in their performance.

\section{Conclusion}\label{conclusion}
In this paper, we examined the effect of quantization methods and pruning methods on deep reinforcement learning algorithms. While the effect depended on the specific DRL algorithm used and the environment in which the agent is trained, the results shared some common patterns. Quantization converts the floating point with 32 bits into an integer with 8 bits model and effectively shrinks the model size while maintaining acceptable performance. We found that PTDQ models generally had the best average return, while PTSQ models might suffer from distribution shifts and had poorer results. DepGraph pruned baseline models by constructing a dependency graph, and we experimented with $L_1$ and $L_2$ pruning. Experiments outlined that $L_2$ pruning was preferred for DRL algorithms on continuous action spaces, and in general, models benefited from $10\%$ $L_2$ pruning with some exceptions. However, while pruning actually removed some neurons, it did not always result in inference speedup or energy saving. 

\bibliographystyle{unsrt}
\bibliography{ref}
\end{document}